\documentclass[sigconf]{acmart}

\usepackage{graphicx}
\usepackage{amsmath}
\usepackage{amsthm}

\usepackage{amssymb}
\usepackage{booktabs} 
\usepackage{epstopdf} 
\usepackage{verbatim} 
\usepackage{subfigure} 
\usepackage{multirow} 
\usepackage{adjustbox} 
\usepackage{xcolor}
\usepackage{array}
\usepackage{enumitem}
\usepackage{bm}

\copyrightyear{2022}
\acmYear{2022}
\setcopyright{acmcopyright}\acmConference[SIGIR '22]{Proceedings of the 45th International ACM SIGIR Conference on Research and Development in Information Retrieval}{July 11--15, 2022}{Madrid, Spain}
\acmBooktitle{Proceedings of the 45th International ACM SIGIR Conference on Research and Development in Information Retrieval (SIGIR '22), July 11--15, 2022, Madrid, Spain}
\acmPrice{15.00}
\acmDOI{10.1145/3477495.3531803}
\acmISBN{978-1-4503-8732-3/22/07}

\begin{document}
\fancyhead{}

\title{A Simple Meta-learning Paradigm for Zero-shot Intent Classification with Mixture Attention Mechanism}

\author{Han Liu}
\affiliation{
\institution{Dalian University of Technology}
\city{Dalian}
\country{China}}
\email{liu.han.dut@gmail.com}

\author{Siyang Zhao}
\affiliation{
\institution{Dalian University of Technology}
\city{Dalian}
\country{China}}
\email{zhao\_siyang@mail.dlut.edu.cn}

\author{Xiaotong Zhang}
\authornote{Corresponding author.}
\affiliation{
\institution{Dalian University of Technology}
\city{Dalian}
\country{China}}
\email{zxt.dut@hotmail.com}

\author{Feng Zhang}
\affiliation{
\institution{Peking University}
\city{Beijing}
\country{China}}
\email{zhangfeng@stu.pku.edu.cn}

\author{Junjie Sun}
\affiliation{
\institution{Dalian University of Technology}
\city{Dalian}
\country{China}}
\email{sunjunjiedlut@hotmail.com}

\author{Hong Yu}
\affiliation{
\institution{Dalian University of Technology}
\city{Dalian}
\country{China}}
\email{hongyu@dlut.edu.cn}

\author{Xianchao Zhang}
\authornotemark[1]
\affiliation{
\institution{Dalian University of Technology}
\city{Dalian}
\country{China}}
\email{xczhang@dlut.edu.cn}

\renewcommand{\shortauthors}{Liu et al.}

\begin{abstract}
Zero-shot intent classification is a vital and challenging task in dialogue systems, which aims to deal with numerous fast-emerging unacquainted intents without annotated training data. To obtain more satisfactory performance, the crucial points lie in two aspects: extracting better utterance features and strengthening the model generalization ability. In this paper, we propose a simple yet effective meta-learning paradigm for zero-shot intent classification. To learn better semantic representations for utterances, we introduce a new mixture attention mechanism, which encodes the pertinent word occurrence patterns by leveraging the distributional signature attention and multi-layer perceptron attention simultaneously. To strengthen the transfer ability of the model from seen classes to unseen classes, we reformulate zero-shot intent classification with a meta-learning strategy, which trains the model by simulating multiple zero-shot classification tasks on seen categories, and promotes the model generalization ability with a meta-adapting procedure on mimic unseen categories. Extensive experiments on two real-world dialogue datasets in different languages show that our model outperforms other strong baselines on both standard and generalized zero-shot intent classification tasks.
\end{abstract}



\maketitle

\section{Introduction}
Dialogue systems are widely used in many real applications, i.e., mobile apps, virtual assistants, smart home and so on \cite{chen2017survey,DBLP:conf/acl/ZhanLLFWL20,DBLP:conf/emnlp/LiuZ0ZZ21,DBLP:conf/sigir/SongYBWWXWY21,DBLP:conf/sigir/XiaXY21}. Understanding user intent is a crucial step in dialogue systems \cite{DBLP:conf/naacl/YuHZDPL21,DBLP:conf/slt/0007KGLAG21,DBLP:conf/icassp/QinLCKZ021,DBLP:conf/sigir/YilmazT20,DBLP:conf/emnlp/QinCLWL19,DBLP:conf/acl/ShenHRJ20}, since its performance will directly affect the downstream decisions and policies. As user interests may change frequently over time and user expression is diverse, many new intents emerge quickly, which motivates zero-shot intent classification. However, related researches are still in infancy, only a few approaches are proposed to tackle this challenge.

\begin{figure*}[t]
	\centering
	\includegraphics[width=0.8\textwidth]{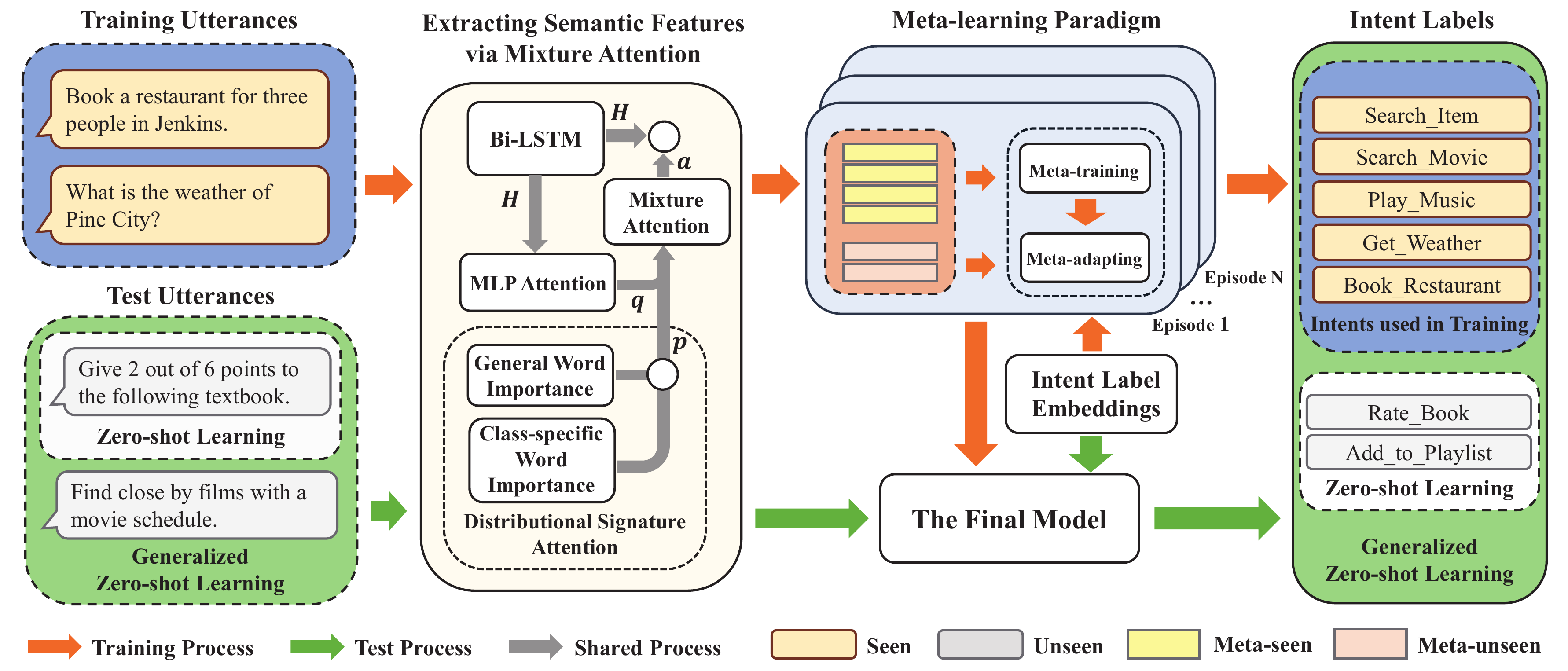}
	\caption{Illustration of the proposed framework. In the training procedure, utterances are first encoded by Bi-LSTM, and then extracted semantic features with the mixture attention mechanism. Specifically, the distributional signature attention utilizes all the seen data to learn the general word importance and leverages the label description information to estimate the class-specific word importance. Meanwhile, the MLP attention automatically calculates different weights for different words. Then the mixture attention, which combines the above two attentions, assigns more appropriate weights for different words. In the meta-learning paradigm, seen data is divided into meta-seen data for meta-training and meta-unseen data for meta-adapting. The model is first trained on meta-seen samples and then is fine-tuned on meta-unseen samples. In the testing procedure, utterances are first encoded by the mixture attention module, and then are classified by the final model which is obtained by the meta-learning paradigm.}
\label{framework}
\end{figure*}

Some works rely on some external resources like label ontologies, manually defined attributes or knowledge graph to find the relationship between seen and unseen intent labels \cite{ferreira2015zero,yazdani2015model,zhang2019integrating,DBLP:journals/corr/abs-2102-02925}. However, such resources are usually unavailable or difficult to obtain, this is because that collecting or producing these resources is labor intensive and time consuming. 

To overcome the above issue, recent works utilize the word embeddings of intent labels instead of the external resources, which are easily to obtained via pretraining on text corpus. Specifically, \citet{baseline-cdssm} and \citet{baseline-zcnn} project the utterances and intent labels to a same semantic space, and then compute the similarities between utterances and intent labels. \citet{xia2018zero} leverage capsule networks \cite{DBLP:conf/nips/SabourFH17} to extract high-level semantic features, and transfer the prediction vectors from seen classes to unseen classes. \citet{liu2019reconstructing} introduce the dimensional attention mechanism to extract semantic features, and then reconstruct the transformation matrices in capsule networks for unseen intents by utilizing abundant latent information of the labeled utterances. \citet{DBLP:conf/acl/YanFLLZWL20} integrate the unknown intent identifier into the model \cite{liu2019reconstructing} to further improve the performance. \citet{DBLP:conf/acl/YeGCCXZWZC20} attempt to use the reinforced self-training framework to learn data selection strategy automatically. \citet{DBLP:conf/ijcai/SiL00LW21} propose a class-transductive framework which uses a multi-task learning objective function to find the inter-intent distinctions and a similarity scorer to associate the inter-intent connections. However, due to the instability in training and the imbalanced classification shift issue, i.e., the model tends to misclassify the unseen test instances into the seen classes, these methods still struggle in generalized zero-shot learning tasks.

After analyzing previous works comprehensively, we find that there are two key points for implementing zero-shot intent classification. The first is to extract better semantic features for utterances, and the second is to strengthen the model generalization ability. Inspired by the success of attention mechanism in representation learning \cite{DBLP:conf/nips/VaswaniSPUJGKP17} and meta-learning strategy in few-shot learning \cite{DBLP:conf/icml/FinnAL17}, we propose a simple yet effective meta-learning paradigm for zero-shot intent classification with mixture attention mechanism. Specifically, we first extract semantic features with the mixture attention mechanism, which can assign more reasonable weights for different words by leveraging the distributional signature attention and the multi-layer perceptron attention simultaneously, thus improving the quality of feature extraction greatly. Then to simulate the multiple zero-shot classification tasks on seen data, we divide seen data into meta-seen data for meta-training and meta-unseen data for meta-adapting, and train the model with a meta-training strategy, which helps to obtain better generalization ability. 

The overall framework of our proposed model is shown in Figure~\ref{framework}. As far as we know, it is the first work to extend meta-learning to the \emph{Zero-Shot Intent Classification} (ZSIC) task. Extensive experimental results on two dialogue datasets in different languages demonstrate that the proposed model dramatically outperforms other baselines on both standard and generalized zero-shot intent classification tasks.

\section{The Proposed Method}
\subsection{Problem Formulation}
Zero-shot learning aims to predict the label $y^*$ of any test sample $\bm x^*$ belonging to an unseen class, leveraging the training data from seen classes. In general, there are two common settings. (1) Standard zero-shot classification: in this case $y^*\in\{C_{\text{unseen}}\}$; (2) Generalized zero-shot classification: in this case $y^*\in\{C_{\text{seen}}, C_{\text{unseen}}\}$. $C_{\text{seen}}$ and $C_{\text{unseen}}$ are the sets of seen and unseen intent classes respectively, and there is no overlap between them, i.e., $C_{\text{seen}} \bigcap C_{\text{unseen}} = \emptyset$.

\subsection{Extracting Semantic Features with Mixture Attention}
Given an utterance with $N$ words $[\bm w_{1},\bm w_{2},...,\bm w_{N}]$, where $\bm w_{t}\in  \mathbb{R}^{d_{w} \times 1}$ is the $t$-th word embedding which can be pre-trained with any language model, we follow \cite{liu2019reconstructing,DBLP:conf/acl/YanFLLZWL20} to use a bidirectional LSTM \cite{DBLP:journals/neco/HochreiterS97} to pre-process these embeddings, i.e., $\bm h_{i}=\text{BiLSTM}(\bm w_{i}) \in \mathbb{R}^{2d_h \times 1}$. Then an utterance can be represented as a matrix $\bm H=[\bm h_1, \bm h_2, \ldots, \bm h_N] \in \mathbb{R}^{2d_h \times N}$, where $d_{h}$ is the hidden dimension of the forward/backward LSTM.

\subsubsection{Distributional Signature Attention} 
Distributional signature attention has shown to be effective in few-shot text classification \cite{DBLP:conf/iclr/BaoWCB20}, which can utilize all the seen data to learn the general word importance and leverage the label description information to estimate the class-specific word importance.

\emph{(1) General word importance.} Frequently occurring words usually have limited discriminative information \cite{DBLP:journals/jd/Jones04}. To downweight frequent words and upweight rare words, we compute general word importance by $s(\bm w_t) = \frac{\varepsilon}{\varepsilon+\text{P}(\bm w_t)}$, where $\varepsilon = 10^{-5}$, and $\text{P}(\bm w_t)$ is unigram likelihood of the $t$-th word over all the seen data.

\emph{(2) Class-specific word importance.} As the importance of a word can be measured by its prediction result for different categories, we can calculate the class-specific word importance by $t(\bm w_{t}) = \mathcal{H}(\text{P}(y|\bm w_{t}))^{-1}$, where $ \mathcal{H}(\cdot)$ is the entropy operator, and $\text{P}(y|\bm w_{t})$ is the conditional probability of $y$ given $\bm w_t$, which can be obtained via a ridge regression classifier. $t(\bm w_t)$ measures the uncertainty of the class label $y$ given the word $\bm w_t$. So words with a skewed distribution will be highly weighted.

In zero-shot scenarios, the label description is the only known information of unseen data. We attempt to use it to calculate the class-specific word importance. Specifically, given the label description embedding matrix $\bm E =[\bm e_1, \bm e_2, \ldots, \bm e_C]^T \in \mathbb{R}^{C \times d_w}$, $\bm e_i$ is the embedding of the $i$-th label description which is computed by the average of its word embeddings, $\bm Y \in \mathbb{R}^{C \times C}$ is the one-hot label matrix, and $C$ is the total number of seen and unseen intent labels. Then we minimize the following regularized squared loss:
\begin{equation}
\mathcal{L} = \|\bm E \bm W-\bm Y\|^{2}_{F}+\beta\|\bm W\|^{2}_{F},
\end{equation}
where $\bm W \in \mathbb{R}^{d_w \times C}$ is the learnable weight matrix, $\|\cdot\|_F$ denotes the Frobenius norm. $\beta$ is the hyperparameter to avoid overfitting and we set $\beta =1$ consistently. Obviously, $\bm W$ has the closed-form solution:
\begin{equation}
\bm W = \bm E^T(\bm E \bm E^T+ \lambda \bm I)^{-1} \bm Y,
\end{equation}
where $\bm I$ is an identity matrix. Then given a word $\bm w_t \in \mathbb{R}^{d_w \times 1}$, we can estimate its conditional probability with:
\begin{equation}
\text{P}(y|\bm w_{t})=\text{softmax}(\bm w_t^T\bm W).
\end{equation}

\emph{(3) Distributional signature attention.} To combine the general and class-specific word importance, we use a bidirectional LSTM to deal with each word, i.e, $\bm z_{t}=\text{BiLSTM}(s(\bm w_{t})||t(\bm w_{t})) \in \mathbb{R}^{2d_b \times 1}$, where $d_b$ is the hidden dimension of the forward/backward LSTM. Then for an utterance, we can calculate the distributional signature attention $\bm p \in \mathbb{R}^{1 \times N}$ by:
\begin{equation}
\bm p = \text{softmax}(\bm F \bm Z),
\end{equation}
where $\bm Z=[\bm z_1, \bm z_2, \ldots, \bm z_N]\in \mathbb{R}^{2d_{b}\times N}$, and $\bm F \in \mathbb{R}^{1\times 2d_{b}}$ is a trainable parameter vector. 

\subsubsection{Multi-layer Perceptron (MLP) Attention} 
MLP attention can automatically calculate the appropriate weights for different words with the guide of loss function \cite{DBLP:conf/aaai/ShenZLJPZ18,DBLP:conf/acl/YanFLLZWL20}. Given an utterance representation $\bm H \in \mathbb{R}^{2d_h \times N}$, the MLP attention $\bm q \in \mathbb{R}^{1 \times N}$can be calculated with:
\begin{equation}
\bm q = \text{softmax}(\bm W_2(\text{ReLU}(\bm W_1 \bm H))),
\end{equation}
where $\bm W_1 \in \mathbb{R}^{d_{a}\times 2d_{h}}$ and $\bm W_2 \in \mathbb{R}^{1 \times d_{a}}$ are trainable parameter matrices.

\subsubsection{Mixture Attention} 
Mixture attention aims to combine the distributional signature attention and MLP attention, and then assigns more appropriate weights for different words in an utterance. We use an adaptive weight vector $\bm b$ to integrate the distributional signature attention and the MLP attention. Specifically, the mixture attention $\bm a \in \mathbb{R}^{1 \times N}$ can be calculated with:
\begin{equation}
\bm a = \bm b [\bm p; \bm q], 
\end{equation}
where $\bm p \in \mathbb{R}^{1 \times N}$ and $\bm q \in \mathbb{R}^{1 \times N}$ are the distributional signature attention and the MLP attention respectively, and $\bm b \in \mathbb{R}^{1\times 2}$ is a trainable parameter vector. Then given an utterance, the final semantic feature $\bm x \in \mathbb{R}^{2 d_h \times 1}$can be computed by:
\begin{equation}
\bm x = \bm H \bm a^T,
\end{equation}
where $\bm a^T$ is the transpose of $\bm a$. Hereinafter, we use $\bm x$ to represent an utterance or an utterance embedding.

\subsection{Meta-learning Paradigm for ZSIC}
Meta-learning has shown promising performance in few-shot learning \cite{DBLP:conf/nips/JeongK20,DBLP:conf/nips/PatacchiolaTCOS20,DBLP:conf/ijcai/SunOZD21,DBLP:conf/ijcai/ZhuLJ20,DBLP:conf/acl/HanFZQGZ21}, while only a few meta-learning based methods are designed for zero-shot learning, which leverage some complex models like VAE or GAN \cite{yu2020episode,verma2020meta}. In addition, these methods mainly concentrate on image domain, and are difficult to directly applied in the intent classification task.

In this work, we propose a simple meta-learning paradigm for \emph{Zero-Shot Intent Classification} (ZSIC), which is trained with the episode strategy. Different from existing meta-learning based few-shot approaches, each episode in our method mimics a zero-shot classification task. In particular, we divide each episode into two phases: \emph{Meta-training} and \emph{Meta-adapting}. Accordingly, to simulate a fake ZSIC task, we randomly divide the seen classes $C_{seen}$ into the meta-seen classes $C_{\text{meta-s}}$ (the samples belonging to $C_{\text{meta-s}}$ are denoted by $X_{\text{meta-s}}$ and used in meta-training phase) and the meta-unseen classes $C_{\text{meta-u}}$ (the samples belonging $C_{\text{meta-u}}$ are denoted by $X_{\text{meta-u}}$ and used in meta-adapting phase). Note that $C_{\text{meta-u}}\cap C_{\text{meta-s}}=\varnothing$.

\subsubsection{Meta-training} Intuitively, the class label description usually contains lots of class-indicative information, so it can be treated as the class prototype (center). If we can learn a model $G$ to project the class label description to the utterance representation space, the embeddings of utterances belonging to the same class should be close to the projection embedding of their label description.

Assume $\bm e_{i} \in \mathbb{R}^{d_w \times 1}$ is the embedding of the $i$-th label description in meta-seen classes, and the model $G$ is a two-layer neural network, we can project $\bm e_i$ to the utterance representation space by:
\begin{equation}\label{LtoU}
G(\bm e_{i})=\text{Tanh}(\bm M_{2}\text{Tanh}(\bm M_{1}\bm e_{i})),
\end{equation}
where $G(\bm e_i) \in \mathbb{R}^{2d_h \times 1}$ denotes the class prototype of the $i$-th category in the utterance representation space. $\bm M_{1} \in \mathbb{R}^{d_s\times d_w}$ and $\bm M_{2} \in \mathbb{R}^{2d_h\times d_s}$ are trainable parameter matrices. Then the probability that an utterance $\bm x$ belongs to the $i$-th category can be calculated by:
\begin{equation}
p_{i}(\bm x)=\frac{exp(-d(\bm x,G(\bm e_{i})))}{\sum_{j}exp(-d(\bm x,G(\bm e_{j})))},
\end{equation}
where $d$ denotes the Euclidean distance. 

In the meta-training procedure, we aim to maximize the probabilities between each utterance and its corresponding class prototype, which can be reformulated by minimizing the following loss function:
\begin{equation}
\mathcal{L}_{train} = -\sum_{\bm x \in X_{\text{meta-s}}}logp_{i}(\bm x). 
\end{equation}

By optimizing $\mathcal{L}_{train}$ with gradient descent, each utterance will be forced to have higher affinity with its corresponding class prototype. During meta-training, we take all utterances in meta-seen classes to train all parameters in the whole model, including mixture attention module and meta-learning module.

\subsubsection{Meta-adapting} To enhance the model generalization ability, we add the meta-adapting phase to refine the model parameters in $G$ using meta-unseen classes. Specifically, given an utterance $\bm x \in X_{\text{meta-u}}$, we use the model learned in meta-training phase to obtain its probability distribution over different classes, i.e., $p(y=i| \bm x)= \frac{exp(-d(\bm x,G(\bm e_{i})))}{\sum_{j}exp(-d(\bm x,G(\bm e_{j})))}$, where $\bm e_{i}$ is the embedding of the $i$-th label description in meta-unseen classes. By minimizing the negative log-probability of the true class $y_k$, i.e.,
\begin{equation}
\mathcal{L}_{adapt}=-\sum_{\bm x \in X_{\text{meta-u}}}log p(y=k|\bm x),
\end{equation}
the model parameters in $G$ can be further updated to adapt the unseen classes. 

\subsubsection{Testing} Given a test utterance $\bm x^*$, its class label $y^*$ can be predicted by: 
\begin{equation}
y^*=\arg \min_k(d(\bm x^*,G(\bm e_{k})),
\end{equation}
where $d$ denotes the Euclidean distance. $\bm e_{k}$ is the embedding of the $k$-th label description. $G(\bm e_{k})$ is the class prototype of the $k$-th category in the utterance representation space.

\section{Experiments}

\begin{table}[t]
	\centering
	\caption{Dataset statistics.}
	\begin{tabular}{l|c|c}
		\toprule
		\textbf{Dataset} & \textbf{SNIPS} & \textbf{SMP} \\
		\midrule
		Vocab Size       &  11641   & 2682  \\
		Number of Samples   &  13802   & 2460   \\
		Average Sentence Length    &  9.05   & 4.86 \\
		Number of Seen Intents     &  5   & 24 \\
        Number of Unseen Intents   &  2  & 6 \\
		\bottomrule
	\end{tabular}
    \label{tab:dataset}
\end{table}

\begin{table*}[t]
    \small
	\centering
	 \caption{Results of generalized zero-shot classification. `Seen', `Unseen' and `Overall' denote the performance on the utterances from seen intents, unseen intents, and both seen and unseen intents respectively.}
	\tabcolsep=0.17cm
	\begin{tabular}{l|cc|cc|cc|cc|cc|cc}
		\toprule
        {} & \multicolumn{6}{c|}{\textbf{SNIPS}} & \multicolumn{6}{c}{\textbf{SMP}}\\
        \cmidrule(lr){2-13}
		{\textbf{Method}} & \multicolumn{2}{c|}{\textbf{Seen}} & \multicolumn{2}{c|}{\textbf{Unseen}} & \multicolumn{2}{c|}{\textbf{Overall}} & \multicolumn{2}{c|}{\textbf{Seen}} & \multicolumn{2}{c|}{\textbf{Unseen}} & \multicolumn{2}{c}{\textbf{Overall}}\\
		\cmidrule(lr){2-3} \cmidrule(lr){4-5} \cmidrule(lr){6-7} \cmidrule(lr){8-9} \cmidrule(lr){10-11} \cmidrule(lr){12-13}
		{} & Acc & F1 & Acc & F1& Acc & F1 & Acc & F1 & Acc & F1& Acc & F1\\
		\midrule
		DeViSE \cite{baseline-devise}       &  0.9481   & 0.6536  & 0.0211   & 0.0398  & 0.4215   & 0.3049  & 0.8040 & 0.6740 & 0.0270 & 0.0310 & 0.5030 & 0.4250 \\
		CMT \cite{baseline-cmt}           &  \textbf{0.9755}   & 0.6648  & 0.0397   & 0.0704  & 0.4438   & 0.3271  & 0.8314 & 0.7221 & 0.0798 & 0.1069 & 0.5398 & 0.4834 \\
		CDSSM \cite{baseline-cdssm}       &  0.9549   & 0.7033  & 0.0111   & 0.0218  & 0.4234   & 0.3194  & 0.6653 & 0.5540 & 0.1436 & 0.1200 & 0.4864 & 0.4052 \\
		ZSDNN \cite{baseline-zcnn}      &  0.9432   & 0.6679  & 0.0682   & 0.1041  & 0.4488   & 0.3493  & 0.7323 & 0.6116 & 0.0590 & 0.0869 & 0.5013 & 0.4316 \\
        IntentCapsNet \cite{xia2018zero}  &  0.9741  & 0.6517 & 0.0000  & 0.0000 & 0.4200  & 0.2810 & \textbf{0.8850}   & 0.7281 & 0.0000  & 0.0000 & 0.5375  & 0.4423  \\
        ReCapsNet \cite{liu2019reconstructing} &  0.9664  & 0.6743 & 0.1121  & 0.1764 & 0.4805  & 0.3911 & 0.8230  & 0.7450 & 0.1720  & 0.1526 & 0.5674  & 0.5124   \\
        RL Self-training \cite{DBLP:conf/acl/YeGCCXZWZC20} &  0.7391  & 0.7558 & 0.5505  & 0.6901 & 0.6257  & 0.7182 & 0.5254  & 0.4791 & 0.4538  & 0.4479 & 0.4654  & 0.4868   \\ 
        CTIR \cite{DBLP:conf/ijcai/SiL00LW21}
        &  0.9693  & 0.6524 & 0.0067  & 0.0132 & 0.4220  & 0.2890 &  0.8322  & 0.7016 & 0.1839  & 0.2083 & 0.5774  & 0.5077  \\
        SEG \cite{DBLP:conf/acl/YanFLLZWL20} &  0.8644  & \textbf{0.8658} & 0.6961  & 0.6931 & 0.7685  & 0.7674 & 0.6821  & 0.7359 & 0.4848  & 0.3806 & 0.6046  & 0.5963   \\
        \midrule
        Ours (w/o gw)     &  0.6328  & 0.5236 & 0.4291  & 0.4671 & 0.5170  & 0.4915 &  0.7584  & 0.7320 & 0.3964  & 0.3942 & 0.6161  & 0.5992 \\
        Ours (w/o cw)   &  0.7514  & 0.6143 & 0.4714  & 0.5512 & 0.5922  & 0.5784 &  0.7383  & 0.7179 & 0.3808  & 0.3051 & 0.5978  & 0.5556\\
		Ours (w/o DS attention)      &  0.7537  & 0.6036 & 0.4671  & 0.5342 & 0.5907  & 0.5641 &  0.7970  & \textbf{0.7464} & 0.2772  & 0.2868 & 0.5927  & 0.5657  \\
		Ours (w/o MLP attention)       &  0.7551  & 0.5944 & 0.4835  & 0.5566 & 0.6006  & 0.5729 &  0.8087  & 0.7446 & 0.2047  & 0.2268 & 0.5713  & 0.5411  \\
        Ours (w/o meta-adapting)     &  0.9507  & 0.6702 & 0.2071  & 0.3267 & 0.5279  & 0.4749 &  0.7651  & 0.7159 & 0.2876  & 0.2916 & 0.5774  & 0.5491 \\
        \midrule
        Ours       &  0.7588  & 0.5944 & 0.5053  & 0.5576 & 0.6146  & 0.5735 &  0.7466  & 0.7268 & 0.4352  & 0.4107 & 0.6242  & 0.6026\\
        Ours (SEG)     &  0.7750  & 0.7903 & \textbf{0.7708}  & \textbf{0.7597} & \textbf{0.7726}  & \textbf{0.7729} &  0.7061  & 0.7046 & \textbf{0.5025}  & \textbf{0.4740} & \textbf{0.6261}  & \textbf{0.6140} \\
		\bottomrule
	\end{tabular}
    \label{tab:result1}
\end{table*}

\subsection{Datasets and Splitting}
\subsubsection{Datasets} We follow \cite{liu2019reconstructing,DBLP:conf/acl/YanFLLZWL20} to evaluate our method on two real dialogue datasets: SNIPS \cite{data_snip} and SMP \cite{data_smp}. SNIPS is an open-source single turn English corpus dialogue, which includes crowdsourced queries distributed among 7 user intents. SMP is a Chinese dialogue corpus for user intent classification released in China National Conference on Social Media Processing. The detailed dataset statistics are summarized in Table \ref{tab:dataset}.

\subsubsection{Data Splitting} For data splitting, we follow \cite{liu2019reconstructing,DBLP:conf/acl/YanFLLZWL20} to divide the datasets. Specifically, for standard zero-shot intent classification, we take all the samples of seen intents as the training set, and all the samples of unseen intents as the test set. For generalized zero-shot intent classification, we randomly take 70\% samples of each seen intent as the training set, and the remaining 30\% samples of each seen intent and all the samples of unseen intents as the test set. 

\subsection{Baselines}
We compare the proposed model with the following state-of-the-art baselines: DeViSE \cite{baseline-devise}, CMT \cite{baseline-cmt}, CDSSM \cite{baseline-cdssm}, ZSDNN \cite{baseline-zcnn}, IntentCapsNet \cite{xia2018zero}, ReCapsNet \cite{liu2019reconstructing},  RL Self-training\cite{DBLP:conf/acl/YeGCCXZWZC20}, CTIR \cite{DBLP:conf/ijcai/SiL00LW21} and SEG \cite{DBLP:conf/acl/YanFLLZWL20}. SEG is a plug-and-play unknown intent detection method, and it integrates with ReCapsNet in the original paper. For fair comparison, we combine our method with SEG and test it in generalized ZSIC. Note that SEG is unsuitable in standard ZSIC as all test utterances are from unseen classes.

In addition, we conduct ablation study to evaluate the contributions of different components in our model. Specifically, we test our model without (w/o) general word importance (gw), class-specific word importance (cw), distributional signature (DS) attention, MLP attention and meta-adapting phase respectively. 

\subsection{Implementation Details}
\subsubsection{Evaluation Metrics}
We adopt two widely used evaluation metrics: accuracy (ACC) and micro-averaged F1 scores (F1) to evaluate the performance. Both metrics are computed with the average value weighted by the support of each class, where the support means the sample ratio of the corresponding class.

\subsubsection{Experiment Settings}
In terms of embeddings, for SNIPS, we use the embeddings pre-trained on English Wikipedia \cite{emb_snip}. For SMP, we use the Chinese word embeddings pre-trained by \cite{emb_smp}. We also evaluate our model with word embeddings pre-trained by BERT \cite{naacl/DevlinCLT19} on the standard zero-shot classification task. In generalized zero-shot tasks, to alleviate the issue that the model tends to classify samples into seen classes, we set a threshold $\lambda$. If a sample's maximum predictive value is less than $\lambda$, the model will classify it among unseen classes. And we set $\lambda=0.6$ on SNIPS, and $\lambda=0.8$ on SMP respectively.

In addition, we use the Adam optimizer \cite{DBLP:journals/corr/KingmaB14} to train the proposed model. For SNIPS, in standard and generalized zero-shot intent classification, we set the initial learning rate 0.006 and 0.002 for meta-training and meta-adapting respectively. And in each episode, we randomly select 4 intents from seen intents as meta-seen classes, and the remaining intents as meta-unseen classes. For SMP, in standard and generalized zero-shot intent classification, we set the initial learning rate 0.008 and 0.004 for meta-training and meta-adapting respectively. And in each episode, we randomly select 21 intents from seen intents as meta-seen classes, and the remaining intents as meta-unseen classes.

\begin{table}[t]
    \small
    \caption{Results of standard zero-shot classification.}
	\centering
	\tabcolsep=0.17cm
	\begin{tabular}{l|cc|cc}
		\toprule
        {\textbf{Method}} & \multicolumn{2}{c|}{\textbf{SNIPS}} & \multicolumn{2}{c}{\textbf{SMP}}\\
        \cmidrule(lr){2-5}
		{} & Acc & F1 & Acc & F1\\
		\midrule
		DeViSE \cite{baseline-devise}       &  0.7447   & 0.7446  & 0.5456   & 0.3875 \\
		CMT \cite{baseline-cmt}          &  0.7396   & 0.7206  & 0.4452   & 0.4245  \\
		CDSSM \cite{baseline-cdssm}      &  0.7588   & 0.7580  & 0.4308   & 0.3765  \\
		ZSDNN \cite{baseline-zcnn}     &  0.7165   & 0.7116  & 0.4615   & 0.3897   \\
        IntentCapsNet \cite{xia2018zero}  &  0.7752  & 0.7750 & 0.4864  & 0.4227 \\
        ReCapsNet \cite{liu2019reconstructing} &  0.7996  & 0.7980 & 0.5418  & 0.4769    \\
        RL Self-training \cite{DBLP:conf/acl/YeGCCXZWZC20} &  0.8253  & 0.8726 & 0.7124  & 0.6587    \\
        CTIR \cite{DBLP:conf/ijcai/SiL00LW21} &  0.6865  & 0.6823 & 0.5462  & 0.5739    \\
        \midrule
        Ours (w/o gw)        &  0.6832  & 0.6676 & 0.5440  & 0.5492 \\
        Ours (w/o cw)        &  0.7349  & 0.7342 & 0.5233  & 0.4460 \\
        Ours (w/o DS attention)        &  0.7480  & 0.7471 & 0.5492  & 0.5292 \\
		Ours (w/o MLP attention)      &  0.8111  & 0.8072 & 0.4741  & 0.4175 \\
        Ours (w/o meta-adapting)       &  0.8390  & 0.8370 & 0.5751  & 0.5436\\
        \midrule
        Ours        &  0.8444  & 0.8443 & 0.6088  & 0.5803 \\
        Ours (BERT)     &  \textbf{0.9139}  & \textbf{0.9135} & \textbf{0.7335}  & \textbf{0.7022} \\
		\bottomrule
	\end{tabular}
    \label{tab:result2}
\end{table}

\subsection{Result Analysis}
Table \ref{tab:result1} and Table \ref{tab:result2} report the experimental results of generalized zero-shot classification and standard zero-shot classification respectively. Some baselines results are taken from \cite{DBLP:conf/acl/YanFLLZWL20}. And the best result is highlighted in bold. Based on the results, we can make the following observations. 

\begin{itemize}
\item In the standard zero-shot intent classification task, our model outperforms other strong baselines on both SNIPS and SMP, which validates the effectiveness of our model in dealing with zero-shot intent classification. In addition, we can also observe that the pre-trained model BERT can improve our model performance effectively. 

\item In the generalized zero-shot intent classification task, our model performs worse than some methods in detecting seen intents. This is because that some baselines tend to classify the test utterances as seen intents, which also explains the reason that these methods perform much worse in detecting unseen intents. From the overall performance, it can be observed that our method can achieve satisfactory results in most cases. 

\item When combining with SEG, our method performs much better than the original version, which indicates that unknown intent detection is helpful to improve generalized zero-shot intent classification and achieve remarkable performance.

\item From ablation study, it can be seen that all modules contribute to the model to some extent, which proves that the mixture attention is powerful and the meta-adapting phase is indispensable.
\end{itemize}

\section{Conclusion}
In this paper, we propose a novel meta-learning paradigm for zero-shot intent classification. The performance gains of the proposed method come from two aspects. By constructing mixture attention mechanism, more reasonable word importance can be learned to obtain better semantic features for utterances. By introducing meta-learning paradigm, the model can achieve better generalization ability to classify utterances in unseen classes. Extensive experimental results confirm the superiority of our method on both standard and generalized zero-shot intent classification tasks. In future work, we plan to extend our framework to deal with multiple-label zero-shot intent detection, and explore more meta-learning strategies for zero-shot intent classification.

\begin{acks}
The authors are grateful to the anonymous reviewers for their valuable comments and suggestions. This work was supported by National Natural Science Foundation of China (No. 62106035, 61876028), and Fundamental Research Funds for the Central Universities (No. DUT20RC(3)040, No. DUT20RC(3)066).
\end{acks}

\bibliography{sample}
\bibliographystyle{ACM-Reference-Format}

\end{document}